\documentclass[lettersize,journal]{IEEEtran}
\usepackage{adjustbox}
\usepackage{amsmath,amsfonts}
\usepackage{algorithmic}
\usepackage{algorithm}
\usepackage{array}
\usepackage{booktabs}
\usepackage[caption=false,font=footnotesize,labelfont=sf,textfont=sf]{subfig}
\usepackage{textcomp}
\usepackage{stfloats}
\usepackage{url}
\usepackage{verbatim}
\usepackage{graphicx}
\usepackage{cite}
\usepackage{threeparttable}
\usepackage{xcolor}
\usepackage{ulem}
\usepackage{multirow}

\begin{document}


\title{SRFlow: A Dataset and Regularization Model for High-Resolution Facial Optical Flow via Splatting Rasterization}


\author{
JiaLin Zhang,~Dong~Li%
\thanks{JiaLin Zhang and Dong Li are with the School of Automation,
Guangdong University of Technology, Guangzhou 510006, China. 
E-mail: 2112304104@gdut.edu.cn, dongli@gdut.edu.cn.}%
}



\maketitle

\begin{abstract}
Facial optical flow supports a wide range of tasks in facial motion analysis. However, the lack of high-resolution facial optical flow datasets has hindered progress in this area. In this paper, we introduce Splatting Rasterization Flow (SRFlow), a high-resolution facial optical flow dataset, and Splatting Rasterization Guided FlowNet (SRFlowNet), a facial optical flow model with tailored regularization losses. These losses constrain flow predictions using masks and gradients computed via difference or Sobel operator. This effectively suppresses high-frequency noise and large-scale errors in texture-less or repetitive-pattern regions, enabling SRFlowNet to be the first model explicitly capable of capturing high-resolution skin motion guided by Gaussian splatting rasterization. Experiments show that training with the SRFlow dataset improves facial optical flow estimation across various optical flow models, reducing end-point error (EPE) by up to 42\% (from 0.5081 to 0.2953). Furthermore, when coupled with the SRFlow dataset, SRFlowNet achieves up to a 48\% improvement in F1-score (from 0.4733 to 0.6947) on a composite of three micro-expression datasets. These results demonstrate the value of advancing both facial optical flow estimation and micro-expression recognition.
\end{abstract}

\begin{IEEEkeywords}
Facial optical flow, facial motion analysis, regularization loss, Gaussian splatting rasterization, micro-expression recognition.
\end{IEEEkeywords}

\section{Introduction}
\IEEEPARstart{F}{acial} optical flow, which quantifies the apparent motion of pixels, surfaces, and edges between consecutive frames, plays a crucial role in capturing subtle, expression-driven facial dynamics \cite{li2025micro}. Characterizing the highly detailed motion patterns underlying facial expressions provides indispensable information for analyzing facial behavior. Automated analysis of facial movements has broad application value, as it provides critical cues for behavior- and cognition-related inference, including emotions, intentions, and predictions \cite{ALLAERT2022434}. A representative application is micro-expression recognition, where involuntary and transient facial muscle movements reveal concealed emotions \cite{ben2021video, liong2016automatic}.

High-resolution facial optical flow datasets are crucial for capturing subtle and highly detailed expression-driven movements, yet they remain scarce due to the difficulty of accurate annotation \cite{lu2024facialflownet}. Most existing optical flow algorithms \cite{teed2020raft, dong2024memflow, Morimitsu2024RecurrentPartialKernel, Morimitsu2025DPFlow} are trained on datasets \cite{kitti2012, butler2012naturalistic, mayer2016large, Dosovitskiy_2015_ICCV, kondermann2016hci} lacking human face structural priors, limiting their performance. Furthermore, high-resolution data introduces additional challenges for facial optical flow estimation \cite{Morimitsu2025DPFlow}, including high-frequency noise and an increased likelihood of prediction errors, which further compromise the accuracy of facial motion analysis.

To address these challenges, this study aims to advance facial-specific optical flow learning by incorporating constraints from both the dataset and the loss function perspectives tailored to facial motion. We also systematically evaluate different optical flow approaches in facial motion analysis to establish stronger baselines and enable more robust applications in detailed facial behavior understanding.

We introduce \textbf{Splatting Rasterization Flow (SRFlow)}, a high-resolution facial optical flow dataset. The Flow Rasterizer, modified from the 3D Gaussian Splatting rasterizer, supports vector transmission between frames. While RGB rendering is performed via the deformation of weighted Gaussian parameters \cite{kerbl3Dgaussians, Wu_2024_CVPR, qian2024gaussianavatars}, these parameters can also be directly transformed to produce optical flow ground truth. We build a dynamic 3D representation trained with GaussianAvatar \cite{qian2024gaussianavatars} under structural constraints, providing detailed, structurally consistent motion information absent in general-purpose datasets. SRFlow dataset is particularly suitable for applications such as micro-expression analysis.


Based on this dataset, we propose \textbf{Splatting Rasterization Guided FlowNet (SRFlowNet)}, an optical flow network designed to address challenges in high-resolution settings, where abundant high-frequency details and large texture-less regions can lead to spurious motion and widespread erroneous predictions. Our design suppresses noise and stabilizes flow estimation, enabling reliable capture of subtle facial motions. SRFlowNet achieves strong performance on the SRFlow test set, and we further validate its effectiveness by applying it to micro-expression recognition with Off-TANet \cite{offtanet}.

In summary, our contributions are as follows:

• We introduce Splatting Rasterization Flow (SRFlow), a high-resolution facial optical flow dataset derived from 3D Gaussian-based dynamic 3D representation, with motion projected onto the image plane using a Flow Rasterizer.

• We present Splatting Rasterization Guided FlowNet (SRFlowNet), the first model incorporating regularization losses to reliably capture subtle facial motion in both textured and texture-less regions, achieving up to a 14\% reduction (from 0.6722 to 0.5791) in F1-ALL for optical flow estimation.

• We demonstrate the effectiveness of SRFlowNet in micro-expression recognition. Trained on the SRFlow dataset, it achieves up to a 48\% improvement in Macro F1-score (from 0.4733 to 0.6947) on the composite dataset.

\section{Related Work}
We first review related work on optical flow estimation, encompassing both traditional and deep learning-based approaches, and highlight key challenges as well as proposed solutions. We then focus on optical flow techniques specifically designed for facial regions, and discuss their applications in downstream tasks such as micro-expression recognition.

\subsection{Optical Flow Estimation}
\label{ss2.1}
Optical flow estimation is fundamentally based on traditional formulations that model the task as a minimization problem balancing data fidelity and smoothness. Classical optical flow models \cite{HORN1981185, tvl1, gf} formulate the task as minimizing an energy functional to estimate the optical flow field. In TVL1 \cite{tvl1}, the energy functional is derived from the brightness constancy assumption, which asserts that the intensity of a point remains constant across consecutive frames. The energy functional consists of a data term and a regularization term. Since the optical flow constraint provides only one equation for two unknowns at each pixel, the system is underdetermined, which motivates the introduction of smoothness regularization to obtain a well-posed estimation.


Building upon traditional optical flow models, subsequent methods have undergone significant architectural evolution. Early optical flow algorithms \cite{Dosovitskiy_2015_ICCV, liu2020learning, yang2019vcn, Yin_2019_CVPR} adopted a coarse-to-fine strategy and relied on an encoder-decoder framework for flow prediction. However, due to insufficient accuracy of flow guidance at coarse levels, traditional coarse-to-fine approaches often fail to capture subtle motions. To address this limitation, optical flow methods based on recurrent network structures \cite{teed2020raft, jiang2021learning, Sun2022SKFlowLearningOptical, xu2022unifying, dong2024memflow, jahedi2024ccmr, wang2024sea, xu2022gmflow} gradually refine and correct predictions by iteratively updating the flow field. More recently, optical flow models employing a recurrent encoder-decoder architecture \cite{Morimitsu2024RAPIDFlowRecurrentAdaptable, Morimitsu2024RecurrentPartialKernel, Morimitsu2025DPFlow} have demonstrated excellent performance. This architecture allows the model to adaptively extract multi-scale features and progressively integrate contextual information through an iterative refinement mechanism.


Although modern architectures improve flow representation, the performance of these models ultimately depends on carefully designed loss functions, which increasingly leverage data-driven supervision facilitated by large-scale labeled datasets. Average end-point error (EPE) is commonly used to evaluate optical flow, and is typically formulated using either the L1 or L2 norm. The L1 norm is robust to outliers but may slow convergence, whereas the L2 norm produces smoother solutions but is sensitive to outliers. Notably, SKFlow \cite{Sun2022SKFlowLearningOptical} employs EPE with the L1 norm to supervise flow estimation. To mitigate limitations of basic EPE losses, several hybrid strategies have been proposed. GMFlowNet \cite{xu2022gmflow} and VCN \cite{yang2019vcn} combine L1 with matching losses, HD³ \cite{Yin_2019_CVPR} employs KL divergence to compare flow distributions, and SEA-RAFT \cite{wang2024sea} and DPFlow \cite{Morimitsu2025DPFlow} adopt mixed Laplacian losses to reduce the influence of ambiguous pixels. ARFlow \cite{liu2020learning} incorporates a smoothness regularization loss based on image gradients, serving as a structural prior to guide unsupervised training in the absence of ground-truth optical flow. However, these regularization strategies are generally not tailored for high-resolution facial motion, where large texture-less regions and fine-grained details introduce additional challenges.


\subsection{Optical Flow Datasets}
\label{ss2.2}
\begin{figure}[!t]
\centering

\subfloat[MPI Sintel \cite{butler2012naturalistic}]{
    \includegraphics[width=0.45\linewidth]{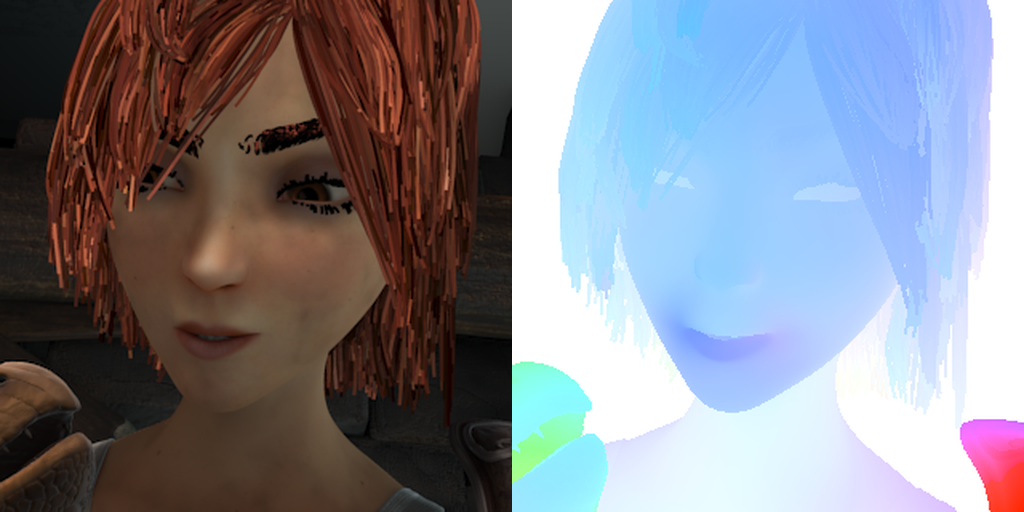}
}
\subfloat[Spring \cite{Mehl2023_Spring}]{
    \includegraphics[width=0.45\linewidth]{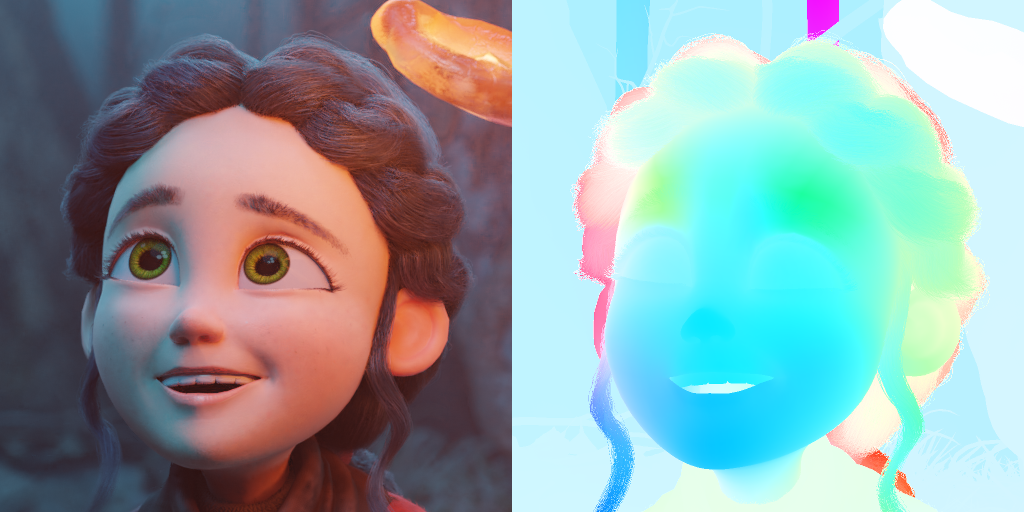}
}

\subfloat[FacialFlowNet \cite{lu2024facialflownet}]{
    \includegraphics[width=0.45\linewidth]{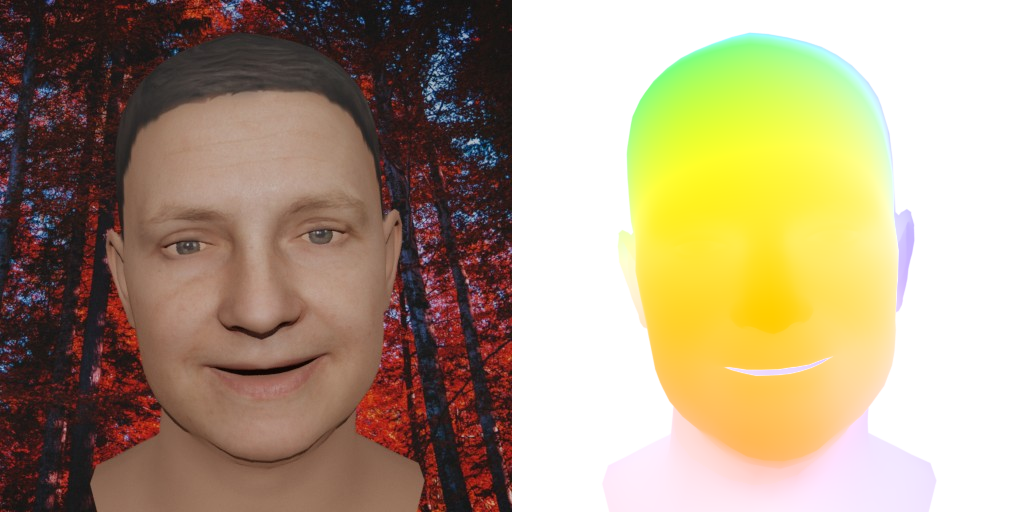}
}
\subfloat[SRFlow]{
    \includegraphics[width=0.45\linewidth]{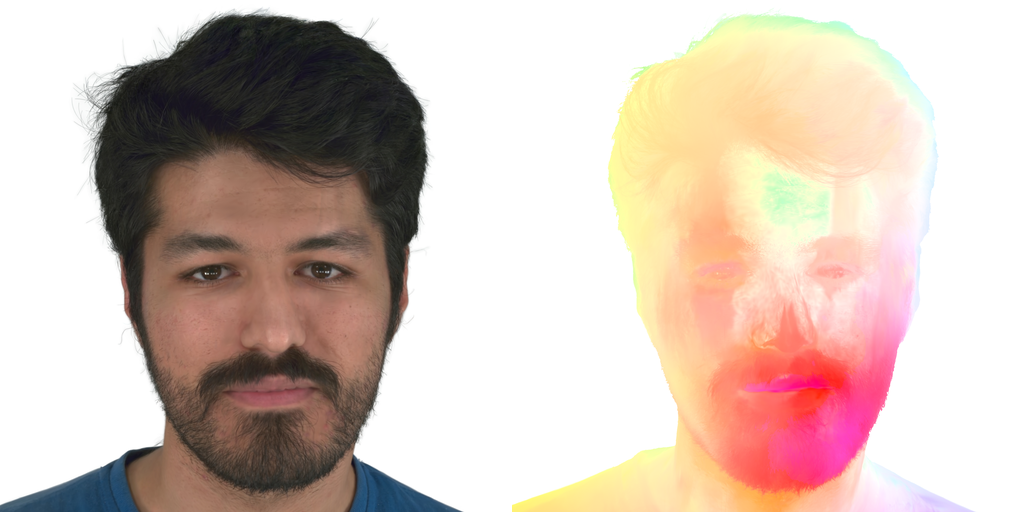}
}

\caption{Illustration of Optical Flow Datasets Featuring Facial Motion}
\label{Fig1}
\end{figure}

Optical flow datasets are generally categorized into real-world datasets \cite{kitti2012, kondermann2016hci} and synthetic datasets \cite{butler2012naturalistic, mayer2016large, Dosovitskiy_2015_ICCV, Mehl2023_Spring}. In real-world datasets, ground truth is typically obtained through LiDAR- or geometry-based reconstruction pipelines. Examples include projecting 3D point clouds in KITTI \cite{kitti2012} and estimating sparse motion in HD1K \cite{kondermann2016hci} using multi-view geometric constraints. However, these datasets primarily target large-scale rigid motions in driving scenarios and consequently provide sparse supervision that does not capture the subtle facial motion required for facial optical flow. Synthetic datasets typically generate ground truth using modified rendering pipelines. While datasets such as FlyingChairs \cite{Dosovitskiy_2015_ICCV} and FlyingThings3D \cite{mayer2016large} do not contain facial motion, other datasets, including MPI Sintel \cite{butler2012naturalistic} and Spring \cite{Mehl2023_Spring}, feature human characters. However, as illustrated in Fig. \ref{Fig1}, these characters are typically simplified cartoon-style avatars that lack realistic facial structures and subtle facial details. As a result, these datasets are not suitable for facial optical flow estimation. To provide a clearer comparison of existing optical flow datasets containing human faces, we summarize their key statistics in Table \ref{table:face_of_datasets}.

Although FacialFlowNet \cite{lu2024facialflownet} was proposed as a dedicated synthetic facial optical flow dataset, its rendering quality is limited by the low fidelity of UV textures and insufficient facial detail. A key limitation is that hair in this dataset is entirely texture-based and lacks volumetric representation. To overcome these limitations, our proposed SRFlow dataset incorporates high-fidelity textures and volumetric hair simulation, and provides a substantially higher rendering resolution of \(2200 \times 3208\), compared with the \(512 \times 512\) resolution of FacialFlowNet. These enhancements enable photorealistic facial motion, providing more reliable supervision for accurate optical flow estimation.

\begin{table}[!t]
\centering
\caption{Comparison of Optical Flow Datasets Containing Human Faces\label{table:face_of_datasets}}
\setlength{\tabcolsep}{4pt}
\begin{tabular*}{\linewidth}{@{\extracolsep{\fill}}l*4c@{}}
\hline
Attribute & MPI Sintel & Spring & FacialFlowNet & SRFlow \\
\hline
Scenes & 35 & 47 & 9635 & 27 \\
Avg. Frames & 47 & 128 & 11 & 413 \\
Total Frames & 1628 & 6000 & 105970 & 11161 \\
Resolution & \(1024\times436\) & \(1920\times1080\) & \(512\times512\) & \(2200\times3208\) \\
\hline
\end{tabular*}
\end{table}

\subsection{Facial Optical Flow}
\label{ss2.3}
Facial optical flow has received increasing attention, with several specialized approaches developed to capture motion in facial regions beyond generic optical flow methods. Alkaddour et al. \cite{ssa} propose a self-supervised model, which we refer to as SSA, that fine-tunes FlowNetS \cite{Dosovitskiy_2015_ICCV} with cyclic losses and additional warping modules. In this work, we use the model variant Exp.~2II described in their paper. While this improves temporal consistency, the approach relies on a general-purpose architecture and does not fully exploit facial motion priors. To leverage facial priors, Lu et al. \cite{lu2024facialflownet} separate expression flow from facial and head motion, enabling more accurate motion decomposition. Zhuang et al. \cite{peng2023facial} introduce differentiable shape priors, including depth, landmarks, and semantic parsing, within a self-supervised non-rigid registration framework, achieving improved robustness under large expressions and head rotations. More recently, Kemmou et al. \cite{kemmou5457671hybrid} propose a hybrid CNN–Transformer GAN to reconstruct missing optical flow caused by occlusions, thereby enhancing downstream facial expression recognition. Despite these advances, existing methods primarily target coarse motion estimation or robustness. Their limited resolution, combined with insufficient attention to subtle facial movements, hinders the accurate capture of motion structures that are essential for high-precision facial optical flow estimation.

After reviewing the related work, we proceed to describe the dataset preparation conducted in this study.

\section{SRFlow Dataset}
Our method is based on a dynamic 3D representation technique that integrates mesh-based modeling with 3D Gaussian Splatting, originally developed for avatar reconstruction and novel view synthesis. By leveraging the high-resolution and multi-view characteristics of the NeRSemble dataset \cite{kirschstein2023nersemble}, we constructed a facial optical flow dataset that captures dense, geometry-consistent motion with sub-pixel spatial precision.

\begin{figure*}[!t]
\centering
\includegraphics[width=1\textwidth]{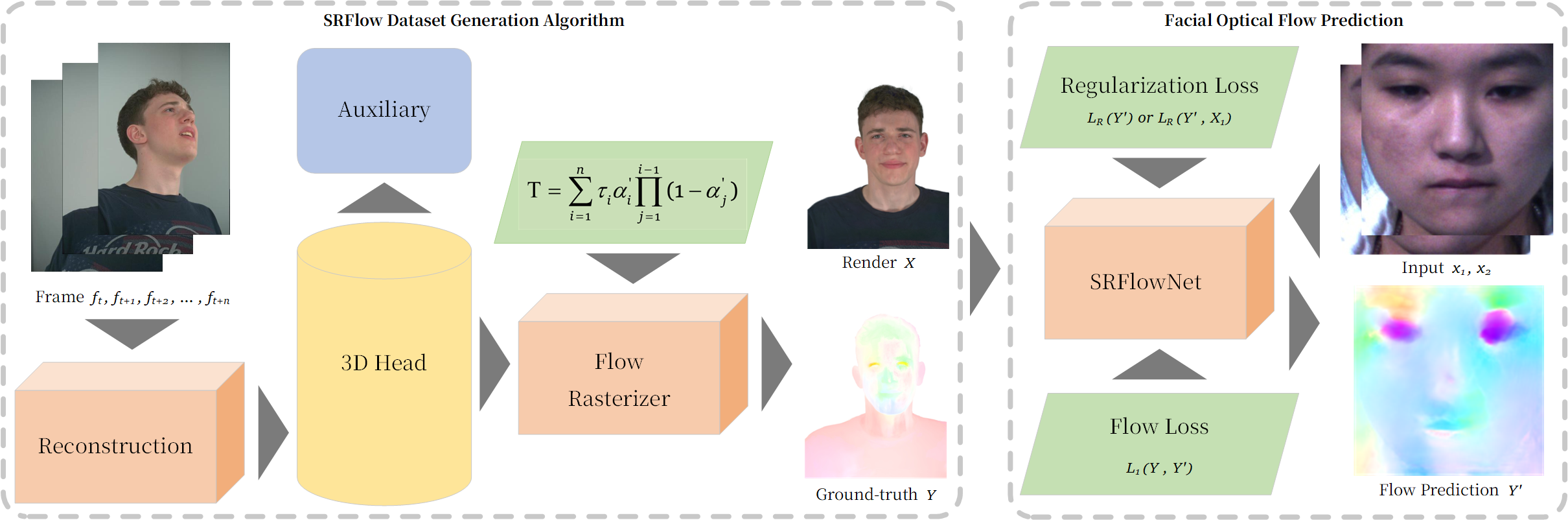}
\caption{Overall Pipeline for Dataset Generation and Network Training}
\label{Fig2}
\end{figure*}

Fig. \ref{Fig2} provides an overview of the pipeline. We first trained a dynamic 3D representation from multi-view video sequences. RGB images, optical flow, and auxiliary motion annotations were then rendered by projecting the 3D representation onto the camera planes. The rendered pairs were subsequently used to supervise facial optical flow networks.


\subsection{3D Gaussian Avatar Reconstruction}
\label{ss3.1}
We reconstructed facial motion using the NeRSemble dataset, which contains more than 4,700 high-frame-rate sequences captured simultaneously by 16 calibrated cameras. The dataset provides time-synchronized multi-view images for 222 subjects, with 157 males and 65 females spanning diverse ages and ethnicities. Each subject contributes 25 motion sequences, including facial expressions, rapid hair movements, emotional variations, speech, and free head motions. To construct our dataset, we selected 27 subjects, ensuring that the chosen motion sequences were as diverse as possible. This dense multi-view coverage, combined with our selection of motion sequences to maximize diversity, enables detailed reconstruction of facial and head motion while providing rich implicit 3D information.


We adopt VHAP \cite{qian2024vhap}, a parametric prior based on FLAME \cite{FLAME:SiggraphAsia2017}, to ensure robust mesh alignment across video frames captured from multiple camera viewpoints. VHAP provides consistent head pose normalization and appearance constraints, generating mesh-guided correspondences even in the absence of predefined landmarks. The aligned outputs from VHAP serve as a reliable initialization for the mesh in GaussianAvatar \cite{qian2024gaussianavatars}.


GaussianAvatar represents each face as a set of continuous 3D Gaussian splats, with each splat encoding a small volumetric region of continuous color and density, and parameterized by the local coordinate frame of its corresponding mesh triangle. This dynamic 3D representation enables smooth and detailed reconstruction of facial geometry and appearance via differentiable rendering. The final color \(C\) of a pixel is computed using alpha blending:

\begin{equation}
     C = \sum_{i=1}^{n} c_i \alpha_i' \prod_{j=1}^{i-1} (1 - \alpha_j').
\end{equation}

where \(c_i \in \mathbb{R}^3\) denotes the color of a Gaussian point represented using spherical harmonics \(SH\), \(\alpha_i'\) is the blending weight computed from the projected area and opacity \(\alpha_i\), and \(\tau_i\) represents the contribution of an individual Gaussian. All Gaussians are depth-sorted prior to blending, and the weighted outputs \(T_i\) determine the final rendered image. Parametric control from FLAME enables a physically consistent representation of head pose and facial motion throughout the sequence.

\subsection{3D Gaussian Dataset Generation}
\label{ss3.2}
We construct a large-scale optical flow dataset using dynamic 3D Gaussian representations generated by GaussianAvatar~\cite{qian2024gaussianavatars}, enabling high-fidelity facial motion modeling. To enhance facial motion rendering, we integrate subject and camera motion by modifying the extrinsic matrix. Each sequence starts and ends with a canonical front-facing camera (camera~9), while intermediate frames apply random rotations on both camera and subject axes. These perturbations introduce rich geometric motion cues and ensure that each sequence contains both canonical views and varied perspectives, providing sufficient diversity to train optical flow networks under realistic facial motion and viewpoint changes.

To generate dense optical flow from the 3D Gaussians, we introduce the Flow Rasterizer, which directly converts the inter-frame motion of Gaussian centers into pixel-wise optical flow. Building upon the original color rendering capability, the Flow Rasterizer additionally accumulates per-Gaussian displacements to produce motion supervision, while preserving geometric consistency and avoiding errors from subsequent 2D interpolation. By operating directly on dynamic Gaussians across consecutive frames, the Flow Rasterizer generates precise and spatially dense optical flow dataset.


Formally, each Gaussian center \(P(X_w, Y_w, Z_w)\) is projected from world coordinates into camera coordinates via the extrinsic matrix:

\begin{equation}
\begin{bmatrix}
X_c \\
Y_c \\
Z_c \\
1
\end{bmatrix}
=
\begin{bmatrix}
R & T \\
\mathbf{0}^T & 1
\end{bmatrix}
\begin{bmatrix}
X_w \\
Y_w \\
Z_w \\
1
\end{bmatrix},
\end{equation}
where \(R \in \mathbb{R}^{3 \times 3}\) and \(T \in \mathbb{R}^{3}\) denote camera rotation and translation. The Flow Rasterizer applies standard perspective projection to map camera coordinates to normalized device coordinates, parameterized by horizontal and vertical field-of-view angles \(\theta_x, \theta_y\):




\begin{equation}
\theta_x = 2 \cdot \arctan(\frac{width}{2 \cdot f_x}), \quad \theta_y = 2 \cdot \arctan(\frac{height}{2 \cdot f_y}), \label{fov}
\end{equation}
where \(f_x, f_y\) are focal lengths and \textit{width}, \textit{height} define the target resolution. A perspective camera is modeled as a view frustum, defined by the distances along the negative Z-axis in camera space to the near and far clipping planes, \(Z_{near}\) and \(Z_{far}\), denoted as \(n\) and \(f\) in the projection matrix. The frustum extents on the near plane, from which the left, right, top, and bottom boundaries \(l, r, t, b\) are defined, are given by:


\begin{equation}
\begin{gathered}
    l = -n \cdot \tan(\theta_x/2), \quad r = n \cdot \tan(\theta_x/2), \\
    b = -n \cdot \tan(\theta_y/2), \quad t = n \cdot \tan(\theta_y/2). \label{frustum_extents}
\end{gathered}
\end{equation}
and the perspective projection matrix is:


\begin{equation}
\begin{bmatrix}
X_{clip} \\
Y_{clip} \\
Z_{clip} \\
W_{clip}
\end{bmatrix}
=
\begin{bmatrix}
\frac{2n}{r-l} & 0 & \frac{r+l}{r-l} & 0 \\
0 & \frac{2n}{t-b} & \frac{t+b}{t-b} & 0 \\
0 & 0 & \frac{f}{f-n} & \frac{-nf}{f-n} \\
0 & 0 & 1 & 0
\end{bmatrix}
\begin{bmatrix}
X_c \\
Y_c \\
Z_c \\
1
\end{bmatrix}.
\end{equation}

After perspective division, the Flow Rasterizer converts normalized device coordinates to pixel coordinates:

\begin{equation}
     u = \frac{(X_{clip}/W_{clip} + 1) \cdot (width - 1)}{2}, \label{u}
\end{equation}

\begin{equation}
    v = \frac{(Y_{clip}/W_{clip} + 1) \cdot (height - 1)}{2}.\label{v}
\end{equation}

Pixel-wise optical flow is computed by the Flow Rasterizer, accumulating projected displacements of each Gaussian across consecutive frames:



\begin{equation}
O_{optical} = 
\begin{bmatrix}
\Delta{U} \\
\Delta{V}
\end{bmatrix}^T
= \sum_{i=1}^{n} 
\begin{bmatrix}
\Delta{u_i} \\
\Delta{v_i}
\end{bmatrix}^T
\alpha_i' \prod_{j=1}^{i-1} (1 - \alpha_j').
\end{equation}

where \((\Delta u_i, \Delta v_i)\) denote the projected motion. This compositing ensures that contributions from all Gaussians are correctly combined, producing dense and accurate optical flow.

In practice, the SRFlow dataset provides all per-pixel annotations as well as intrinsic and extrinsic camera matrices. The total number of image pairs in the dataset is 11161, and these were partitioned into 6791, 1212, and 3158 pairs for training, validation, and test sets, respectively.

\section{SRFlowNet Model}
We propose SRFlowNet, a splatting rasterization–guided optical flow model for high-resolution facial motion learning. We first benchmark multiple pretrained optical flow models on the SRFlow dataset and assess their downstream performance on micro-expression recognition to select the most suitable backbone. Building on this backbone, we introduce lightweight regularization losses that enhance structural consistency, suppress high-frequency artifacts, and preserve subtle facial motions.


\subsection{Model Backbone}\label{ss4.1}
The choice of backbone architecture plays a critical role in determining the quality of optical flow estimation for subtle and short-term facial motion. Unlike generic optical flow benchmarks that primarily involve large displacements and rigid motion, facial motion reconstruction in micro-expression analysis demands high sensitivity to subtle, low-amplitude, and spatially localized deformations. To identify a suitable backbone for SRFlowNet under this challenging setting, we systematically benchmark a diverse set of representative pretrained optical flow models on the SRFlow dataset, and further evaluate their retrained versions in both optical flow accuracy and downstream micro-expression recognition performance. This two-stage evaluation allows us to assess not only raw flow estimation capability but also task relevance, leading to a principled selection of the final backbone architecture.

To determine the backbone for SRFlowNet, we first benchmark several representative pretrained optical flow models on the SRFlow dataset. All models are trained using a widely adopted combination of standard optical flow datasets, including FlyingChairs \cite{Dosovitskiy_2015_ICCV}, FlyingThings3D \cite{mayer2016large}, Sintel \cite{butler2012naturalistic}, KITTI \cite{kitti2012}, and HD1K \cite{kondermann2016hci}. This multi-dataset pretraining strategy is commonly used in optical flow research and is known to provide strong cross-dataset generalization ability.

During evaluation, we use the SRFlow test set and crop each frame to a resolution of \(1600 \times 1200\) to focus on facial regions and eliminate background interference. The quantitative results are summarized in Table~\ref{table:pretrained_results}. For clarity, we highlight the top three results for each metric as follows: the best result in bold blue, the second in blue, and the third with underlined text. Based on these pretrained results, we select four candidate backbones for further investigation: MemFlow \cite{dong2024memflow}, SKFlow \cite{Sun2022SKFlowLearningOptical}, DPFlow \cite{Morimitsu2025DPFlow}, and RPKNet \cite{Morimitsu2024RecurrentPartialKernel}. These models consistently rank among the top performers across multiple metrics.

\begin{table}[!t]
\caption{Comparison of Optical Flow Methods Using Pretrained Models\label{table:pretrained_results}}
\centering
\setlength{\tabcolsep}{2pt}
\begin{tabular}{*7c}
\toprule
        Methods&EPE ↓& px1 ↑&px3 ↑& px5 ↑&F1-ALL ↓&WAUC ↑\\
        \midrule
     RAFT\cite{teed2020raft} &0.5494 &0.8416 &0.9671 &0.9925 &3.2936 &79.8585\\
     GMA\cite{jiang2021learning} &0.5424 &0.8416 &0.9676 &0.9928 &3.2420 &80.0873\\
     SKFlow\cite{Sun2022SKFlowLearningOptical} * &0.5361 &0.8452 &0.9678 &0.9927 &3.2159 &80.3533\\
     Unimatch\cite{xu2022unifying} &0.7176 &0.7730 &0.9626 &0.9915 &3.7419 &73.1524\\
     RPKNet\cite{Morimitsu2024RecurrentPartialKernel} * &\textcolor{blue}{0.5115} &\textcolor{blue}{0.8548} &\uline{0.9692} &\uline{0.9930} &\uline{3.0814} &\textcolor{blue}{\textbf{84.6435}}\\
     MemFlow\cite{dong2024memflow} * &\textcolor{blue}{\textbf{0.5081}} &\textcolor{blue}{\textbf{0.8551}} &\textcolor{blue}{\textbf{0.9699}} &\textcolor{blue}{0.9932} &\textcolor{blue}{\textbf{3.0071}} &80.3933\\
     SEA-RAFT\cite{wang2024sea} &0.5314 &0.8451 &0.9676 &0.9927 &3.2396 &\uline{80.5733}\\
     CCMR+\cite{jahedi2024ccmr} &0.5867 &0.8225 &0.9624 &0.9914 &3.7565 &78.7038\\
     DPFlow\cite{Morimitsu2025DPFlow} * &\uline{0.5233} &\uline{0.8486} &\textcolor{blue}{0.9695} &\textcolor{blue}{\textbf{0.9933}} &\textcolor{blue}{3.0453} &\textcolor{blue}{84.1902}\\
     \bottomrule
\end{tabular}
\begin{tablenotes}
\item We select four models as backbone candidates and mark them with an asterisk (*) superscript.
\end{tablenotes}
\end{table}

To further assess domain adaptability, we retrain all four selected models on the SRFlow training set using the training configurations specified in their original papers, except that we reduce the learning rate by half to better accommodate the high-resolution imagery and subtle motion patterns in the SRFlow dataset. The retrained results, reported in Table~\ref{table:retrained_results}, show consistent improvements across most metrics. This confirms that facial motion reconstruction exhibits strong domain-specific characteristics that are not fully captured by models trained solely on generic optical flow datasets.

Beyond optical flow accuracy, we further evaluate each retrained model on a downstream micro-expression recognition task on a composite dataset consisting of SAMM~\cite{davison2016samm}, CASME II~\cite{yan2014casme}, and SMIC~\cite{li2013spontaneous}, following the protocol described in Section~\ref{ss5.1.2}, with results reported in Table~\ref{tab:retrained_micro_exp_performance}. Although retrained MemFlow and DPFlow achieve higher optical flow accuracy on the SRFlow test set, retrained SKFlow obtains higher scores on four out of five metrics on the micro-expression recognition task. Optical flow benchmarks evaluate pixel-wise displacement error over the full image domain, whereas micro-expression recognition relies on classification accuracy computed from motion-derived features within localized facial regions. Additionally, optical flow training is performed on consecutive frame pairs, whereas micro-expression recognition computes flow between onset and apex frames, reflecting different temporal sampling strategies between training and evaluation.


Based on this two-stage evaluation, we select SKFlow as the backbone of SRFlowNet for subsequent experiments.


\begin{table*}[!t]
\caption{Performance of Retrained Optical Flow Models on the SRFlow Dataset\label{table:retrained_results}}
\centering
\begin{threeparttable}
\begin{tabular*}{\linewidth}{@{\extracolsep{\fill}}c*6c*6c@{}}
\toprule
& \multicolumn{6}{c}{SRFlow} 
& \multicolumn{6}{c}{Improvement} \\
\cmidrule(lr){2-7} \cmidrule(lr){8-13}
Methods & EPE ↓ & px1 ↑ & px3 ↑ & px5 ↑ & F1-ALL ↓ & WAUC ↑
& EPE & px1 & px3 & px5 & F1-ALL & WAUC \\
\midrule

SKFlow+SRFlow
& \uline{0.3998} & \uline{0.8975} & \uline{0.9933} & \uline{0.9992} 
& \uline{0.6722} & 83.8335
& 25.42\% & 6.19\% & 2.63\% & 0.65\% & 79.10\% & 4.33\% \\

RPKNet+SRFlow
& 0.4655 & 0.8728 & 0.9820 & 0.9967 & 1.8001 & \uline{85.2226}
& 8.99\% & 2.11\% & 1.32\% & 0.37\% & 41.58\% & 0.68\% \\

MemFlow+SRFlow
& \textcolor{blue}{\textbf{0.2953}} & \textcolor{blue}{\textbf{0.9374}} & \textcolor{blue}{\textbf{0.9965}} & \textcolor{blue}{\textbf{0.9998}} 
& \textcolor{blue}{\textbf{0.3502}} & \textcolor{blue}{86.9671}
& 41.88\% & 9.62\% & 2.74\% & 0.66\% & 88.35\% & 8.18\% \\

DPFlow+SRFlow
& \textcolor{blue}{0.3348} & \textcolor{blue}{0.9244} & \textcolor{blue}{0.9960} & \textcolor{blue}{0.9996}
& \textcolor{blue}{0.3961} & \textcolor{blue}{\textbf{88.8041}}
& 36.02\% & 8.93\% & 2.73\% & 0.63\% & 86.99\% & 5.48\% \\
\bottomrule
\end{tabular*}
\end{threeparttable}
\end{table*}

\begin{table}[!t]
    \centering
    \caption{Performance of Retrained Optical Flow Models on the Composite Dataset for Micro-Expression Recognition}
    \label{tab:retrained_micro_exp_performance}
    \begin{tabular*}{0.48\textwidth}{@{\extracolsep{\fill}} l c c c c c @{}}
        \toprule
        Methods & $P_M$ & $R_M$ & $F_{1M}$ & $F_{1\mu}$ & $G_M$ \\
        \midrule
        SKFlow+SRFlow & \textcolor{blue}{\textbf{0.7013}} & \textcolor{blue}{\textbf{0.7016}} & \textcolor{blue}{\textbf{0.6845}} & \textcolor{blue}{\textbf{0.7736}} & \uline{0.5083} \\
        RPKNet+SRFlow & \uline{0.6886} & \uline{0.6850} & \uline{0.6696} & \uline{0.7551} & \textcolor{blue}{\textbf{0.5523}} \\
        MemFlow+SRFlow & 0.6494 & 0.6358 & 0.6254 & 0.7212 & 0.4466 \\     
        DPFlow+SRFlow & \textcolor{blue}{\textbf{0.7013}} & \textcolor{blue}{0.7005} & \textcolor{blue}{0.6797} & \textcolor{blue}{0.7577} & \textcolor{blue}{0.5317} \\
        \bottomrule
    \end{tabular*}
\end{table}

\subsection{Regularization Loss for High-Resolution Facial Optical Flow}
\label{ss4.2}
This section introduces a set of regularization losses designed to improve high-resolution facial optical flow estimation on the SRFlow dataset. In high-resolution facial imagery, rich high-frequency components such as fine skin textures and illumination variations introduce complex local gradients, while texture-less or repetitive regions often provide insufficient motion constraints, both of which make optical flow estimation more challenging for subtle facial motion. Without additional constraints, these factors can cause the model to overfit to local appearance variations or propagate unreliable motion estimates across regions, motivating the need for regularization to suppress noise-like responses and encourage spatial coherence. To address these challenges, we incorporate four lightweight regularization losses: Total Variation Regularization Loss \ref{ss4.2.1}, Flow Difference Regularization Loss \ref{ss4.2.2}, Mean Image Gradient Activation Regularization Loss \ref{ss4.2.3}, and Image Gradient Variance Activation Regularization Loss \ref{ss4.2.4}. These regularizations stabilize motion estimation and encourage alignment with meaningful facial structures, enhancing robustness in high-resolution scenarios and benefiting downstream micro-expression recognition across datasets. To validate their effectiveness, we train SRFlowNet on the SRFlow dataset using different regularization losses under the same training settings described in Section~\ref{ss5.1.1}. The full training configuration is detailed below.


\subsubsection{Total Variation Regularization Loss (TVR)}
\label{ss4.2.1}
We adopt a TVL1-inspired smoothness regularization to compute spatial gradients of the predicted optical flow. The predicted flow is decomposed into horizontal and vertical components, and gradients are computed independently for each channel in both horizontal and vertical directions.

\begin{equation}
    R(c) = \frac{1}{H \times W} \sum_{x,y} \big( |\nabla_x c(x, y)| + |\nabla_y c(x, y)| \big).
\end{equation}

where \(R(c)\) measures the total variation of a single flow channel \(c\), which can be the horizontal component \(u\) or the vertical component \(v\). \(H\) and \(W\) denote the height and width of the flow field, respectively. The horizontal and vertical gradients \(\nabla_x\) and \(\nabla_y\) are implemented using normalized Sobel operators, obtained by dividing the standard \(3\times3\) Sobel kernels by 8. This normalization prevents over-penalizing large gradient magnitudes introduced by the weighted structure of the Sobel operator.

\begin{equation}
    L_{TVR} = \lambda_{N} \sum_{i=0}^{n-1} \gamma^{n-i-1} \big( R(u^i) + R(v^i) \big).
\end{equation}

Here, the regularization weight is \(\lambda_N = 0.05\), \(n\) denotes the total number of prediction stages, and the decay factor \(\gamma \in (0,1)\) assigns progressively smaller weights to earlier-stage predictions.




\subsubsection{Flow Difference Regularization Loss (FDR)}
\label{ss4.2.2}
Following ARFlow \cite{liu2020learning}, we extend the TVR by replacing the normalized Sobel operator with flow difference and incorporating a mask \(M_{bg}\) to focus on valid flow regions.

\begin{equation}
\begin{gathered}
    D_x(f) = \frac{f(x, y+stride) - f(x, y)}{stride}, \\
    D_y(f) = \frac{f(x+stride, y) - f(x, y)}{stride}.
\end{gathered}
\end{equation}

\(D_x\) and \(D_y\) compute flow difference of the optical flow along the horizontal and vertical directions. The stride, which determines the pixel spacing for the flow difference, is set to 1.

\begin{equation}
\begin{gathered}
    L_{FDR} = \lambda_N \sum_{i=0}^{n-1} \gamma^{n-i-1} \Big( D_x(f^i) \odot M_{bg}^x \\
    + D_y(f^i) \odot M_{bg}^y \Big).
\end{gathered}
\end{equation}

Specifically, \(M_{bg}^x\) and \(M_{bg}^y\) are obtained by removing the last column and the last row of \(M_{bg}\), respectively, to match the spatial dimensions of \(D_x(f)\) and \(D_y(f)\). We observe that when using flow difference, optical flow metrics improve, as shown in Table \ref{table:ablation_results}, but micro-expression recognition performance decreases, as reported in Table \ref{tab:micro_exp_full_performance}. Flow difference compute approximate horizontal and vertical derivatives using only immediate neighbors, which are strictly aligned with the two flow channels and thus more consistent with the representation of optical flow directions. As a result, FDR based on flow difference enforces simpler and axis-aligned regularization, which tends to suppress subtle local variations and may discard fine facial motion details, as shown in Fig.~\ref{Fig3}. In contrast, the Sobel operator aggregates information from a larger spatial neighborhood, including diagonal directions introduced by the corner weights. This spatial coupling allows Sobel-based regularization to better preserve subtle motion patterns and local structural variations, although the resulting gradients involve mixtures of neighboring directions.

\subsubsection{Mean Image Gradient Activation Regularization Loss (MIGAR)}
\label{ss4.2.3}
ARFlow computes per-pixel weights for the regularization loss based on the gradients of two input frames at consecutive time steps, at time \(t\) and \(t+1\). Considering optical flow as the per-pixel displacement from the first frame to the second frame, the spatial structure is largely determined by the first frame, while the magnitude and direction depend on both frames. Motivated by this observation, MIGAR constrains flow structure using only the gradients of the first frame. Unlike ARFlow, which uses a fixed exponential base, MIGAR adopts an adaptive exponential weighting scheme based on the mean gradient magnitude of the first frame. Inspired by FDR, MIGAR applies the standard Sobel operator instead of flow difference. Per-pixel weights are applied, so normalization of the Sobel operator is unnecessary. 

\begin{equation}
    G_{I_1} = \frac{1}{3}\sum_{c=1}^3 
    \sqrt{
        \left(\nabla_x I_1^{c}(x,y)\right)^2 +
        \left(\nabla_y I_1^{c}(x,y)\right)^2
    }
\end{equation}

\(G_{I_1}\) represents the gradient magnitude of the first frame \(I_1\), averaged over the three color channels using the Sobel operator to capture spatial structure.

\begin{equation}
    base = \exp \left\{ \frac{1}{H \times W} \sum_{x,y} G_{I_1}(x,y) \right\}.
\end{equation}

The variable \(base\) is the adaptive exponential base, computed as the exponent of the mean gradient magnitude over all pixels, which serves as a global normalization factor for per-pixel weighting.

\begin{equation}
    w = base^{-G_{I_1}(x,y)}.
\end{equation}

The per-pixel weight \(w(x,y)\) is strictly positive due to exponential parameterization, and it amplifies the contrast between low- and high-gradient regions.

\begin{equation}
\begin{aligned}
    M_{total} &= M_{bg}(x,y) \odot \big( (\nabla_x M_{bg}(x,y) > 0) \\
    &\qquad \wedge (\nabla_y M_{bg}(x,y) > 0) \big).
\end{aligned}
\end{equation}

The mask \(M_{total}\) is constructed by combining the background mask \(M_{bg}\) with its gradients.

\begin{equation}
\begin{aligned}
R_{wpp}(c) 
&= \frac{1}{H \times W} \sum_{x,y} 
w(x,y) \Big(
\big| \nabla_x \big( c(x,y) \odot M_{total}(x,y) \big) \big| \\
&\qquad\qquad + 
\big| \nabla_y \big( c(x,y) \odot M_{total}(x,y) \big) \big|
\Big).
\end{aligned}
\end{equation}

The weighted per-pixel regularization \(R_{wpp}(c)\) computes horizontal and vertical gradients of the masked single optical flow channel \(c\) and scales them by per-pixel weights \(w(x,y)\). The total loss function is described below:

\begin{equation}
    L_{MIGAR} = \sum_{i=0}^{n-1} \gamma^{n-i-1} \left( R_{wpp}(u^i) + R_{wpp}(v^i) \right).
\end{equation}

\subsubsection{Image Gradient Variance Activation Regularization Loss (IGVAR)}
\label{ss4.2.4}
In the original MIGAR, the exponential weighting term uses the mean gradient of the entire image as its base. Background regions contribute gradients that do not correspond to meaningful facial motion. To address this issue, IGVAR replaces the base with the variance of gradient magnitudes computed only within valid regions. In contrast to the mean gradient used in MIGAR, the variance better reflects local gradient dispersion and emphasizes structural differences within facial regions. Formally, the base is defined as:

\begin{equation}
\begin{aligned}
\Omega &= \left\{ (x,y) \mid M_{bg}(x,y) = 1 \right\}, \\
base &= \max \left( \frac{\mathrm{Var}\left( \{ G_{I_1}(x,y) \mid (x,y) \in \Omega \} \right)}{100}, e \right).
\end{aligned}
\end{equation}

Dividing by 100 keeps the base in a range comparable to the original MIGAR, and the lower bound \(e\) ensures numerical stability during exponentiation. All other terms remain unchanged compared with MIGAR. This adjustment serves a role similar to the global coefficient \(\alpha\) in ARFlow while avoiding excessively small weights in low-texture regions. By replacing only the base of the exponential weighting term with the variance of gradients within valid regions, IGVAR preserves the structure of MIGAR while making the per-pixel weights more robust to valid regions.

\section{Results and Discussion}
We evaluate SRFlowNet and the proposed regularization designs through a comprehensive set of analyses aimed at understanding both optical flow quality and its downstream impact. We begin with ablation experiments that evaluate each loss design on optical flow accuracy. We then conduct a qualitative visualization study on the SRFlow test set and three micro-expression datasets, comparing the predicted flow fields of different models to evaluate their ability to capture subtle and spatially localized facial motions. Finally, we measure downstream micro-expression recognition performance to determine how improvements in flow fidelity translate into better high-level semantic understanding. Together, these results provide an integrated view of how regularized optical flow estimation influences both low-level motion quality and the reliability of micro-expression analysis.


\subsection{Experimental Setup}\label{ss5.1}
This section describes the experimental settings used for both optical flow estimation and micro-expression recognition. We first introduce the training details of SRFlowNet and other optical flow models, followed by the setup for the downstream micro-expression recognition task.

\subsubsection{Optical Flow Estimation Setup}\label{ss5.1.1}
All optical flow models were trained under a unified experimental configuration. The SRFlow training set comprises 6{,}789 image pairs with an original resolution of \(2200 \times 3208\). Data augmentation was applied consistently across all methods. Training was conducted on two RTX A6000 Ada GPUs with a total of 96 GB memory. The batch size was set to 8. All models were trained for 45 epochs.

During training, input images were randomly cropped to obtain patches of size \(800 \times 512\). The learning rate was set to \(1.25 \times 10^{-4}\) for all models.


\subsubsection{Micro-Expression Recognition Setup}\label{ss5.1.2}
To evaluate the effectiveness of optical flow features for micro-expression recognition, we conducted a comprehensive comparison and further analyzed several variants of the proposed SRFlowNet to assess the impact of different loss designs on performance.


We conduct micro-expression recognition experiments on three benchmark datasets: SAMM, CASME II, and SMIC. For SAMM and CASME II, we directly adopt the officially provided onset and apex frame annotations, whereas for SMIC, which does not include prelabeled onset and apex frames, we use the predictions generated by CapsuleNet~\cite{Quang2019Capsulenet}. Optical flow is computed between the onset and apex frames and transformed into optical strain of size \(112 \times 112\) that served as flow features input to Off-TANet~\cite{offtanet}, a lightweight triplet-attention-based classifier. Detailed statistics for SAMM, CASME II, SMIC, and the composite dataset are listed in Table~\ref{tab:me_datasets_statistics}~\cite{ben2021video}.


\begin{table}[!t]
    \centering
    \caption{Statistics of SAMM, CASME II, and SMIC for Micro-Expression Recognition}
    \label{tab:me_datasets_statistics}
    \setlength{\tabcolsep}{3pt}
    \begin{tabular}{l c c c c c c}
         \toprule
         & \multicolumn{3}{c}{No. of Samples per Class} & \multicolumn{2}{c}{Gender} & \multirow{2}{*}{Ethnicities} \\
         \cmidrule(lr){2-4} \cmidrule(lr){5-6}
         Dataset & Positive & Negative & Surprise & Male & Female & \\
         \midrule
         SAMM \cite{davison2016samm} & 26 & 92 & 15 & 12 & 16 & 13 \\
         CASME\ II \cite{yan2014casme} & 32 & 88 & 25 & 10 & 14 & 1 \\
         SMIC \cite{li2013spontaneous} & 51 & 65 & 41 & 10 & 6 & 2 \\
         Composite & 109 & 245 & 81 & 32 & 36 & 13 \\
         \bottomrule
    \end{tabular}
\end{table}

The learning rate for the micro-expression recognition is set to \(6.5 \times 10^{-4}\). Following the Off-TANet evaluation protocol, metrics are recorded every 20 epochs, and the final score for each metric is defined as the maximum cross-validation average across all recorded epochs.


\subsection{Ablation of Regularization Loss Designs}
\label{ss5.2}
To assess the contribution of the proposed loss designs, we compare different SRFlowNet variants that are built upon the SKFlow architecture and trained with different loss functions against both pretrained and retrained SKFlow baselines. Results are reported in Table \ref{table:ablation_results}. 




\begin{table}
\caption{Optical Flow Performance in the Ablation Studies\label{table:ablation_results}}
\centering
\setlength{\tabcolsep}{2pt}
\begin{tabular*}{0.48\textwidth}{@{\extracolsep{\fill}}*7c@{}}
\toprule
        Methods& EPE ↓& px1 ↑&px3 ↑& px5 ↑&F1-ALL ↓&WAUC ↑\\
        \midrule
     SKFlow\cite{Sun2022SKFlowLearningOptical} &0.5361 &0.8452 &0.9678 &0.9927 &3.2159 &80.3533\\
     SKFlow+SRFlow & \uline{0.3998} & 0.8975 & \textcolor{blue}{0.9933} & \textcolor{blue}{0.9992} & \textcolor{blue}{0.6722} & 83.8335 \\
     SRFlowNet-TVR & 0.4056 &0.8977 &0.9924 &0.9988 &0.7622 &83.6409\\
     SRFlowNet-FDR &\textcolor{blue}{\textbf{0.3946}} &\textcolor{blue}{0.8991} &\uline{0.9932} &\textcolor{blue}{\textbf{0.9992}} &\uline{0.6815} &\textcolor{blue}{\textbf{84.0652}}\\
     SRFlowNet-MIGAR &\textcolor{blue}{0.3955} &\uline{0.8987} &\textcolor{blue}{\textbf{0.9942}} &\textcolor{blue}{\textbf{0.9993}} &\textcolor{blue}{\textbf{0.5791}} &\textcolor{blue}{83.9568}\\
     SRFlowNet-IGVAR &0.4000 &\textcolor{blue}{\textbf{0.8993}} &0.9927 &\uline{0.9989} &0.7270 &\uline{83.8422}\\
     \bottomrule
\end{tabular*}
\end{table}

Although these losses share the common goal of stabilizing high-resolution facial flow by suppressing noisy or implausible motion, they operate through slightly different smoothing mechanisms. Consistent with their shared nature, all four regularizers improve over the retrained SKFlow baseline, though their relative strengths show only minor variations across metrics. SRFlowNet-FDR achieves the lowest EPE and highest WAUC, SRFlowNet-MIGAR obtains the lowest F1-ALL, and SRFlowNet-IGVAR yields competitive performance overall. These differences, while modest, reflect how each regularizer biases the optimization toward a slightly different balance between smoothness and detail preservation. Notably, while most variants follow similar performance trends, SRFlowNet-TVR behaves as an exception and performs less favorably because it is the only regularizer that does not incorporate a background mask, causing its smoothing term to be applied in regions without valid motion. Except for SRFlowNet-FDR, the other regularizers use the Sobel operator for gradient computation. While effective at suppressing noise, Sobel operator also thickens edges and may introduce spurious boundaries, as illustrated in Fig. \ref{Fig3}. These factors collectively explain why SRFlowNet-FDR produces the most stable and coherent flow fields, whereas other variants may generate spurious edges, resulting in the superior performance of SRFlowNet-FDR across optical flow metrics.



Because the four losses target similar failure modes and differ only in their smoothing behavior, no single loss dominates universally. This is further reflected in the downstream micro-expression recognition experiments as shown in Section \ref{ss5.4}, where different datasets favor different regularization schemes, suggesting mild dataset-dependent preferences rather than fundamentally different behaviors.

We do not include a variant that combines any of these losses. As these losses are all regularizers, stacking them would largely amplify the same smoothing effect, increasing the risk of oversuppression of subtle motions—the signals critical for micro-expression analysis. For clarity and interpretability, we isolate each regularizer’s effect individually. Exploring combined-loss configurations is possible future work but is not central to our study.

\subsection{Qualitative Result of Facial Optical Flow}
\label{ss5.3}
\begin{figure*}[!t]
\centering
\includegraphics[width=1\textwidth]{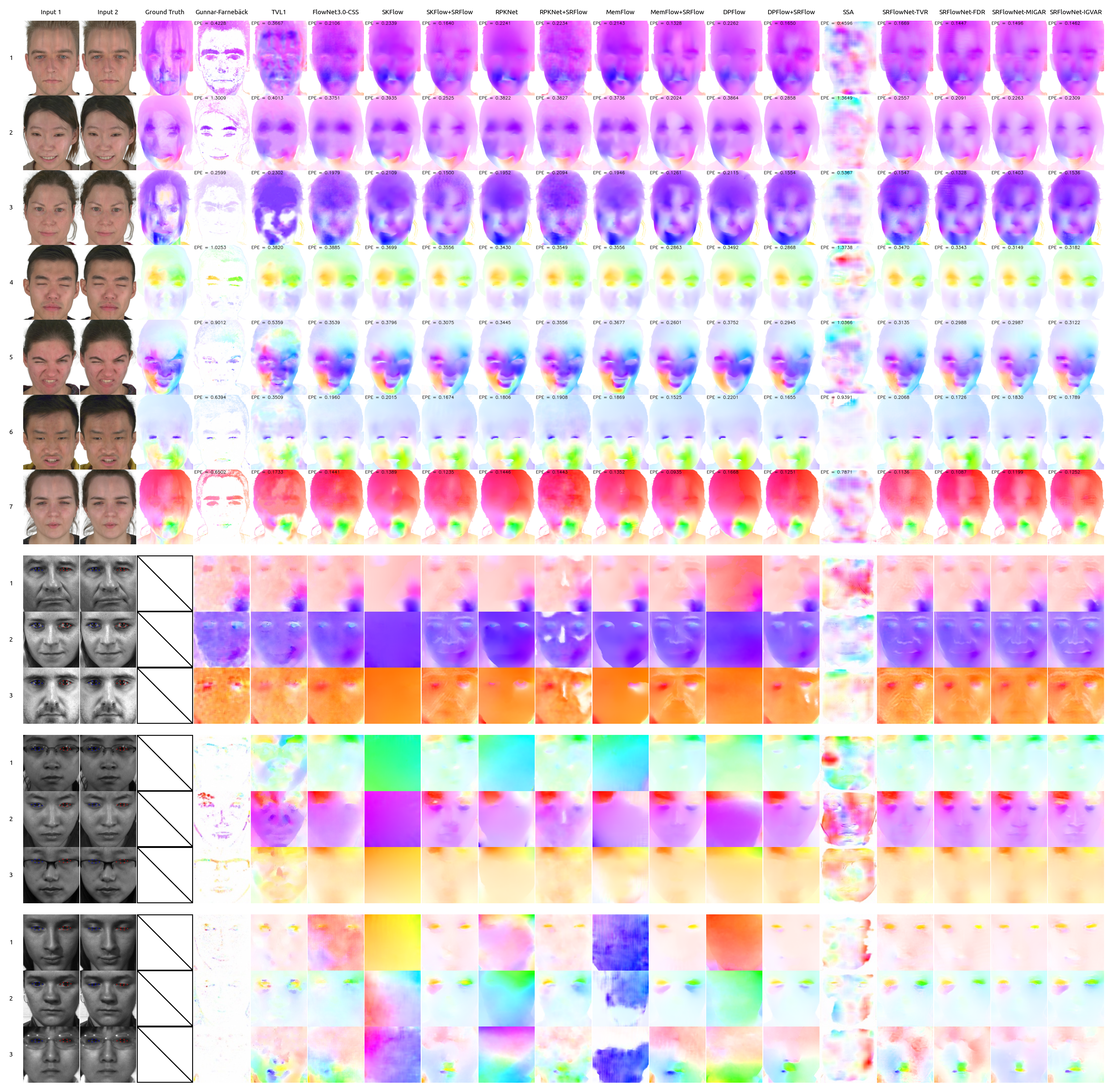}
\caption{Qualitative Visualization of Optical Flow Estimations Across All Evaluated Networks.}
\label{Fig3}
\end{figure*}

In this section, we present a qualitative evaluation of facial optical flow. Fig. \ref{Fig3} shows representative examples from the SRFlow test set. The ground-truth flow is provided by the dataset, while the predictions are generated by our evaluated networks. We also present visual results computed from the onset and apex frames of three micro-expression datasets: SAMM \cite{davison2016samm}, CASME II \cite{yan2014casme}, and SMIC HS \cite{li2013spontaneous}. Although these observations are based on a few qualitative examples, they provide valuable insights that complement the quantitative results reported in Tables \ref{table:pretrained_results}, \ref{table:retrained_results}, and \ref{table:ablation_results}.

Differences in optical flow predictions are clearly visible across networks on the SRFlow test set. Among traditional methods, Gunnar-Farneb{\"a}ck produces the sparsest outputs, whereas TV-L1 yields denser flow fields but still lacks fine structural details. For deep learning models, FlowNet3.0-CSS provides predictions that closely match the ground truth, although slightly blurrier. Pretrained models including SKFlow, RPKNet, MemFlow, and DPFlow generally align with the ground truth. Retraining these models on SRFlow further improves alignment in local facial regions such as the forehead, eyebrows, eyes, and mouth, producing smoother and more coherent motion fields without exaggerating amplitudes. Facial optical model SSA produces coarser predictions with missing details and occasional directional errors, which can be critical for micro-expression analysis. Among the SRFlowNet variants, SRFlowNet-MIGAR and SRFlowNet-FDR generate smoother and more coherent motion patterns, particularly in the chin region, whereas SRFlowNet-TVR and SRFlowNet-IGVAR still show partial false edges in some areas.

Qualitative results on the micro-expression datasets show consistent patterns. Among traditional methods, Gunnar-Farneb{\"a}ck produces the sparsest outputs, whereas TVL1 yields denser but overly smooth flow fields. Among deep learning models, FlowNet3.0-CSS provides predictions closer to the ground truth but becomes noticeably blurry on challenging image pairs. Pretrained SKFlow and pretrained DPFlow lose significant facial details. In contrast, pretrained RPKNet and pretrained MemFlow preserve eye motion on SAMM and CASME II, but degrade on the lower-quality SMIC frames. Retraining these models on SRFlow improves motion coherence across datasets. However, retrained DPFlow and retrained RPKNet still fail to capture lip deformation on SAMM, and retrained SKFlow misses some global motion on CASME II. Facial optical model SSA generates sparse or hole-like flow fields. Among the SRFlowNet variants, SRFlowNet-FDR and SRFlowNet-MIGAR produce the most stable and coherent flow fields, whereas SRFlowNet-TVR and SRFlowNet-IGVAR introduce more false edges, highlighting the impact of their respective regularization losses.

Models trained on the SRFlow dataset achieve the strongest facial optical flow performance, consistently capturing subtle motion patterns in local facial regions. This demonstrates the value of our high-resolution facial motion dataset, whose spatial detail, accurate annotations, and carefully controlled generation pipeline provide supervision that preserves fine local deformations often lost in lower-quality datasets. Consequently, networks trained on SRFlow are more sensitive to small-amplitude, localized facial motions, reinforcing their suitability for learning detailed facial dynamics.

\subsection{Micro-Expression Recognition Results}
\label{ss5.4}
We evaluate micro-expression recognition across multiple datasets using various optical flow methods.

On the SAMM dataset, as shown in Table~\ref{tab:samm_performance}, SRFlowNet-TVR achieves the strongest overall performance, ranking first in three out of five evaluation metrics. The traditional methods Gunnar–Farneb{\"a}ck and TVL1 remain competitive, each ranking among the top three methods across multiple metrics, suggesting that robust low-level motion modeling remains beneficial on SAMM. Nevertheless, retrained MemFlow and retrained RPKNet do not consistently outperform their pretrained versions, with several metrics showing slight degradation. This behavior contrasts with the general trend observed on other datasets, where retraining on SRFlow typically leads to improved performance. Although SAMM includes a larger and more ethnically diverse set of subjects, as summarized in Table~\ref{tab:me_datasets_statistics}, the number of samples per subject is sparse and uneven. This imbalance introduces strong cross-subject variability, which can limit the effectiveness of large-scale retraining.

\begin{table}[!t]
    \centering
    \caption{Performance of Optical Flow Methods on SAMM}
    \label{tab:samm_performance}
    \begin{tabular}{l c c c c c}
        \toprule
        Methods & $P_M$ & $R_M$ & $F_{1M}$ & $F_{1\mu}$ & $G_M$ \\
        \midrule
        Gunnar-Farneb{\"a}ck \cite{gf} & \textcolor{blue}{0.6187} & \textcolor{blue}{\textbf{0.6786}} & \textcolor{blue}{0.6124} & \textcolor{blue}{\textbf{0.7429}} & \textcolor{blue}{0.4722} \\
        TVL1 \cite{tvl1} & \underline{0.6086} & \textcolor{blue}{0.6594} & 0.5995 & \underline{0.7278} & 0.4538 \\
        FlowNet3.0-CSS \cite{ilg2018occlusions} & \underline{0.6086} & 0.6431 & 0.6014 & 0.7137 & 0.4627 \\
        SKFlow \cite{Sun2022SKFlowLearningOptical} & 0.5799 & 0.6458 & 0.5856 & 0.7129 & 0.4286 \\
        SKFlow+SRFlow & 0.5899 & 0.6522 & 0.5966 & 0.7224 & 0.4316 \\
        RPKNet \cite{Morimitsu2024RecurrentPartialKernel} & 0.5964 & 0.6475 & 0.5965 & 0.7156 & 0.4520 \\
        RPKNet+SRFlow & 0.5990 & 0.6445 & 0.5939 & 0.7277 & 0.4668 \\
        MemFlow \cite{dong2024memflow} & 0.6012 & 0.6452 & 0.5928 & 0.7105 & 0.4477 \\
        MemFlow+SRFlow & 0.5889 & \underline{0.6534} & 0.5920 & 0.7205 & 0.4286 \\
        DPFlow \cite{Morimitsu2025DPFlow} & 0.5557 & 0.6333 & 0.5791 & 0.6948 & 0.4286 \\
        DPFlow+SRFlow & 0.5996 & 0.6475 & 0.5953 & 0.7199 & \underline{0.4673} \\
        SSA \cite{ssa} & 0.5814 & \textcolor{blue}{\textbf{0.6786}} & \textcolor{blue}{0.6124} & \textcolor{blue}{\textbf{0.7429}} & 0.4286 \\
        SRFlowNet-TVR & \textcolor{blue}{\textbf{0.6261}} & 0.6404 & \textcolor{blue}{\textbf{0.6156}} & \textcolor{blue}{0.7384} & \textcolor{blue}{\textbf{0.4725}} \\
        SRFlowNet-FDR & 0.5964 & 0.6486 & 0.5949 & 0.7150 & 0.4538 \\
        SRFlowNet-MIGAR & 0.5989 & 0.6517 & \underline{0.6029} & 0.7206 & 0.4538 \\
        SRFlowNet-IGVAR & 0.5886 & 0.6475 & 0.5889 & 0.7150 & 0.4421 \\
        \bottomrule
    \end{tabular}
\end{table}

Results on CASME II, as shown in Table~\ref{tab:casme2_performance}, clearly demonstrate the superiority of our approach. SRFlowNet-TVR ranks first in all five evaluation metrics, outperforming all competing methods by a clear margin. In addition, FlowNet3.0-CSS and SRFlowNet-FDR consistently rank among the top three across most metrics. For SRFlowNet-FDR, this highlights the benefits of our motion regularization and high-resolution facial supervision, whereas FlowNet3.0-CSS occasionally benefits from its ability to capture coarse or large-scale motions. Moreover, all pretrained models benefit substantially from retraining on SRFlow, indicating that CASME II is particularly sensitive to refined motion modeling and regularization strategies. This can be attributed to the fact that CASME II offers the highest-quality recordings, which aligns well with the high-resolution and high-fidelity facial motion supervision provided by the SRFlow dataset. Overall, the characteristics of CASME II further amplify the benefits of both the SRFlow dataset and the regularization integrated within SRFlowNet.

\begin{table}[!t]
    \centering
    \caption{Performance of Optical Flow Methods on CASME II}
    \label{tab:casme2_performance}
    \begin{tabular}{l c c c c c}
        \toprule
        Methods & $P_M$ & $R_M$ & $F_{1M}$ & $F_{1\mu}$ & $G_M$ \\
        \midrule
        Gunnar-Farneb{\"a}ck \cite{gf} & 0.6351 & 0.6374 & 0.6123 & 0.7238 & 0.4776 \\
        TVL1 \cite{tvl1} & 0.7410 & 0.7118 & 0.6932 & 0.7610 & 0.5486 \\
        FlowNet3.0-CSS \cite{ilg2018occlusions} & \textcolor{blue}{0.7853} & \textcolor{blue}{0.7722} & \textcolor{blue}{0.7709} & \textcolor{blue}{0.8415} & \textcolor{blue}{0.6741} \\
        SKFlow \cite{Sun2022SKFlowLearningOptical} & 0.4712 & 0.5405 & 0.4699 & 0.6168 & 0.2917 \\
        SKFlow+SRFlow & 0.7770 & 0.7431 & 0.7429 & 0.8042 & 0.5917 \\
        RPKNet \cite{Morimitsu2024RecurrentPartialKernel} & 0.6243 & 0.6269 & 0.5987 & 0.6945 & 0.4196 \\
        RPKNet+SRFlow & 0.7536 & 0.7293 & 0.7220 & 0.7962 & 0.5391 \\
        MemFlow \cite{dong2024memflow} & 0.6733 & 0.6251 & 0.6248 & 0.7442 & 0.3930 \\
        MemFlow+SRFlow & 0.7300 & 0.7141 & 0.7080 & 0.8066 & 0.5387 \\
        DPFlow \cite{Morimitsu2025DPFlow} & 0.6292 & 0.6010 & 0.5821 & 0.6811 & 0.4072 \\
        DPFlow+SRFlow & 0.7674 & 0.7522 & \underline{0.7468} & 0.8175 & 0.5957 \\
        SSA \cite{ssa} & 0.6318 & 0.5855 & 0.5753 & 0.6459 & 0.3923 \\
        SRFlowNet-TVR & \textcolor{blue}{\textbf{0.7914}} & \textcolor{blue}{\textbf{0.7937}} & \textcolor{blue}{\textbf{0.7831}} & \textcolor{blue}{\textbf{0.8426}} & \textcolor{blue}{\textbf{0.6826}} \\
        SRFlowNet-FDR & \underline{0.7823} & \underline{0.7534} & 0.7454 & \underline{0.8194} & 0.6165 \\
        SRFlowNet-MIGAR & 0.7635 & 0.7352 & 0.7295 & 0.7824 & \underline{0.6218} \\
        SRFlowNet-IGVAR & 0.7573 & 0.7164 & 0.7124 & 0.7919 & 0.5585 \\
        \bottomrule
    \end{tabular}
\end{table}

On the SMIC dataset, as shown in Table~\ref{tab:smic_performance}, despite having the lowest recording quality among the three micro-expression datasets, SRFlowNet-TVR achieves the best overall performance. FlowNet3.0-CSS and TVL1 obtain competitive results and occasionally outperform our method on one or two individual metrics. This is mainly because they are more effective at capturing coarse or large-scale motions, which can better align with certain evaluation criteria under degraded imaging conditions. In contrast, SRFlow provides high-resolution facial supervision that is better suited for modeling subtle facial motions, explaining why our method performs best overall while being only slightly surpassed on isolated metrics in low-quality scenarios. Nevertheless, SRFlowNet-TVR still ranks first in two out of the five metrics and remains among the top three in most others, demonstrating strong stability and robustness under low-resolution conditions.


\begin{table}[!t]
    \centering
    \caption{Performance of Optical Flow Methods on SMIC}
    \label{tab:smic_performance}
    \begin{tabular}{l c c c c c}
        \toprule
        Methods & $P_M$ & $R_M$ & $F_{1M}$ & $F_{1\mu}$ & $G_M$ \\
        \midrule
        Gunnar-Farneb{\"a}ck \cite{gf} & 0.4226 & 0.4058 & 0.3663 & 0.4239 & 0.2436 \\
        TVL1 \cite{tvl1} & \underline{0.5868} & \textcolor{blue}{0.5766} & \underline{0.5612} & \underline{0.6325} & \textcolor{blue}{\textbf{0.4283}} \\
        FlowNet3.0-CSS \cite{ilg2018occlusions} & \textcolor{blue}{0.5947} & \textcolor{blue}{\textbf{0.5862}} & \textcolor{blue}{0.5640} & \textcolor{blue}{\textbf{0.6615}} & 0.3342 \\
        SKFlow \cite{Sun2022SKFlowLearningOptical} & 0.2787 & 0.3022 & 0.2585 & 0.3345 & 0.1110 \\
        SKFlow+SRFlow & 0.5352 & 0.4973 & 0.4932 & 0.6028 & 0.2813 \\
        RPKNet \cite{Morimitsu2024RecurrentPartialKernel} & 0.4897 & 0.4864 & 0.4373 & 0.5107 & 0.3154 \\
        RPKNet+SRFlow & 0.5443 & 0.5181 & 0.5083 & 0.6173 & 0.3337 \\
        MemFlow \cite{dong2024memflow} & 0.4852 & 0.4274 & 0.4127 & 0.4775 & 0.2530 \\
        MemFlow+SRFlow & 0.5730 & 0.5620 & 0.5456 & 0.6314 & 0.3422 \\
        DPFlow \cite{Morimitsu2025DPFlow} & 0.3755 & 0.3291 & 0.3279 & 0.4087 & 0.2342 \\
        DPFlow+SRFlow & 0.5705 & 0.5438 & 0.5356 & 0.6241 & 0.3019 \\
        SSA \cite{ssa} & 0.4271 & 0.4042 & 0.3657 & 0.4235 & 0.1829 \\
        SRFlowNet-TVR & \textcolor{blue}{\textbf{0.6009}} & \underline{0.5749} & \textcolor{blue}{\textbf{0.5654}} & \textcolor{blue}{0.6542} & \underline{0.3741} \\
        SRFlowNet-FDR & 0.5034 & 0.5032 & 0.4808 & 0.5834 & 0.3290 \\
        SRFlowNet-MIGAR & 0.5539 & 0.5512 & 0.5207 & 0.6132 & 0.3354 \\
        SRFlowNet-IGVAR & 0.5763 & 0.5719 & 0.5313 & 0.6111 & \textcolor{blue}{0.3961} \\
        \bottomrule
    \end{tabular}
\end{table}



Results on the composite Dataset, as shown in Table~\ref{tab:micro_exp_full_performance}, highlight the overall effectiveness of our approach across diverse micro-expression scenarios. SRFlowNet-TVR and SRFlowNet-IGVAR both achieve two first-place rankings across evaluation metrics. TVR demonstrates higher absolute performance in individual metrics with two second-place results, whereas IGVAR achieves top-three rankings across all metrics, showing greater overall stability, highlighting how improvements in optical flow fidelity and smoothness lead to enhanced micro-expression recognition and more reliable high-level semantic interpretation. Overall, the results suggest that combining multiple datasets amplifies the advantages of SRFlow and the regularization integrated within SRFlowNet, enabling consistent and robust performance across varied micro-expression conditions.

\begin{table}[!t]
    \centering
    \caption{Performance of Optical Flow Methods on the Composite Dataset}
    \label{tab:micro_exp_full_performance}
    \begin{tabular}{l c c c c c}
        \toprule
        Methods & $P_M$ & $R_M$ & $F_{1M}$ & $F_{1\mu}$ & $G_M$ \\
        \midrule
        Gunnar-Farneb{\"a}ck \cite{gf} & 0.5707 & 0.5765 & 0.5510 & 0.6617 & 0.3802 \\
        TVL1 \cite{tvl1} & \textcolor{blue}{0.7118} & \underline{0.7038} & \underline{0.6904} & 0.7675 & \textcolor{blue}{0.5442} \\
        FlowNet3.0-CSS \cite{ilg2018occlusions} & 0.6932 & 0.6788 & 0.6656 & 0.7492 & 0.5116 \\
        SKFlow \cite{Sun2022SKFlowLearningOptical} & 0.4650 & 0.5285 & 0.4733 & 0.5906 & 0.3181 \\
        SKFlow+SRFlow & 0.7013 & 0.7016 & 0.6845 & \underline{0.7736} & 0.5083 \\
        RPKNet \cite{Morimitsu2024RecurrentPartialKernel} & 0.5668 & 0.5726 & 0.5413 & 0.6566 & 0.3643 \\
        RPKNet+SRFlow & 0.6886 & 0.6850 & 0.6696 & 0.7551 & \textcolor{blue}{\textbf{0.5523}} \\
        MemFlow \cite{dong2024memflow} & 0.5695 & 0.6174 & 0.5733 & 0.6793 & 0.3926 \\
        MemFlow+SRFlow & 0.6494 & 0.6358 & 0.6254 & 0.7212 & 0.4466 \\
        DPFlow \cite{Morimitsu2025DPFlow} & 0.6022 & 0.6084 & 0.5798 & 0.6725 & 0.4388 \\        
        DPFlow+SRFlow & 0.7013 & 0.7005 & 0.6797 & 0.7577 & 0.5317 \\
        SSA \cite{ssa} & 0.5099 & 0.5186 & 0.4880 & 0.6019 & 0.3105 \\
        SRFlowNet-TVR & \textcolor{blue}{\textbf{0.7177}} & \textcolor{blue}{0.7059} & \textcolor{blue}{\textbf{0.6947}} & \textcolor{blue}{0.7781} & 0.5402 \\
        SRFlowNet-FDR & 0.6634 & 0.6529 & 0.6388 & 0.7208 & 0.4801 \\
        SRFlowNet-MIGAR & 0.6956 & 0.6905 & 0.6706 & 0.7477 & 0.5329 \\
        SRFlowNet-IGVAR & \underline{0.7095} & \textcolor{blue}{\textbf{0.7070}} & \textcolor{blue}{0.6912} & \textcolor{blue}{\textbf{0.7792}} & \underline{0.5412} \\
        \bottomrule
    \end{tabular}
\end{table}


Different optical flow algorithms exhibit varying performance across micro-expression datasets due to intrinsic differences in dataset characteristics and algorithm design. Such variations affect the ability of each algorithm to capture subtle facial motion accurately. Traditional methods typically produce blocky motion estimates, and the strong regularization term they impose further oversmooths the flow, suppressing subtle facial expressions. In contrast, deep learning models adapt to the distributions of general optical flow training data but may remain sensitive to low-texture regions or dataset-specific noise patterns. Our SRFlowNet variants incorporate regularization losses specifically designed to stabilize high-resolution facial flow and emphasize meaningful motion regions. Although these losses generally enhance robustness, their relative benefit varies with each dataset’s characteristics and class imbalance. Evaluating multiple metrics therefore provides a comprehensive assessment that reflects both overall and minority-class performance.



\section{Conclusion and Future Work}
In this paper, we presented a framework for facial optical flow estimation guided by Gaussian splatting rasterization. We constructed \textbf{SRFlow}, a high-resolution facial optical flow dataset, which provides the foundation for training \textbf{SRFlowNet}. The proposed network incorporates tailored regularization losses that exploit facial masks and image gradients, computed via flow difference or Sobel operator, to suppress high-frequency noise and large-scale artifacts in texture-less or repetitive regions. Experimental results demonstrate that both the SRFlow dataset and the SRFlowNet architecture significantly advance facial optical flow estimation and lead to notable improvements in micro-expression recognition. SRFlowNet achieves state-of-the-art performance on one metric in SAMM, four in CASME II, two in SMIC, and all five in the composite dataset, validating the benefit of jointly improving facial optical flow estimation and its downstream applications.

For further investigation and improvement, future work related to this study may include exploring the creation of high-resolution micro-expression datasets and high-resolution recognition networks. While our work shows that high-resolution facial optical flow can substantially benefit micro-expression recognition, current datasets for both optical flow and micro-expression tasks remain limited in resolution. As a result, most micro-expression recognition models are trained and evaluated on relatively low-resolution inputs. Future work will focus on creating high-resolution micro-expression datasets and developing high-resolution micro-expression recognition networks, which would fully exploit detailed optical flow information and further improve recognition performance.

\bibliographystyle{IEEEtran} 
\bibliography{IEEEabrv,refs} 

\begin{thebibliography}{10}
\providecommand{\url}[1]{#1}
\csname url@samestyle\endcsname
\providecommand{\newblock}{\relax}
\providecommand{\bibinfo}[2]{#2}
\providecommand{\BIBentrySTDinterwordspacing}{\spaceskip=0pt\relax}
\providecommand{\BIBentryALTinterwordstretchfactor}{4}
\providecommand{\BIBentryALTinterwordspacing}{\spaceskip=\fontdimen2\font plus
\BIBentryALTinterwordstretchfactor\fontdimen3\font minus \fontdimen4\font\relax}
\providecommand{\BIBforeignlanguage}[2]{{%
\expandafter\ifx\csname l@#1\endcsname\relax
\typeout{** WARNING: IEEEtran.bst: No hyphenation pattern has been}%
\typeout{** loaded for the language `#1'. Using the pattern for}%
\typeout{** the default language instead.}%
\else
\language=\csname l@#1\endcsname
\fi
#2}}
\providecommand{\BIBdecl}{\relax}
\BIBdecl

\bibitem{li2025micro}
W.~Li, H.~Lin, Y.~Yan, and Z.~Feng, ``Micro-expression recognition based on the fusion of facial landmark points and optical flow features,'' \emph{Journal of Electronic Imaging}, vol.~34, no.~1, pp. 013\,028--013\,028, 2025.

\bibitem{ALLAERT2022434}
\BIBentryALTinterwordspacing
B.~Allaert, I.~Ward, I.~Bilasco, C.~Djeraba, and M.~Bennamoun, ``A comparative study on optical flow for facial expression analysis,'' \emph{Neurocomputing}, vol. 500, pp. 434--448, 2022. [Online]. Available: \url{https://www.sciencedirect.com/science/article/pii/S0925231222006610}
\BIBentrySTDinterwordspacing

\bibitem{ben2021video}
X.~Ben, Y.~Ren, J.~Zhang, S.-J. Wang, K.~Kpalma, W.~Meng, and Y.-J. Liu, ``Video-based facial micro-expression analysis: A survey of datasets, features and algorithms,'' \emph{IEEE transactions on pattern analysis and machine intelligence}, vol.~44, no.~9, pp. 5826--5846, 2021.

\bibitem{liong2016automatic}
S.-T. Liong, J.~See, K.~Wong, and R.~C.-W. Phan, ``Automatic micro-expression recognition from long video using a single spotted apex,'' in \emph{Asian conference on computer vision}.\hskip 1em plus 0.5em minus 0.4em\relax Springer, 2016, pp. 345--360.

\bibitem{lu2024facialflownet}
J.~Lu, R.~He, S.~Zhou, W.~Tan, and B.~Yan, ``Facialflownet: Advancing facial optical flow estimation with a diverse dataset and a decomposed model,'' in \emph{Proceedings of the 32nd ACM International Conference on Multimedia}, ser. MM '24.\hskip 1em plus 0.5em minus 0.4em\relax New York, NY, USA: Association for Computing Machinery, 2024, p. 2194–2203.

\bibitem{teed2020raft}
Z.~Teed and J.~Deng, ``Raft: Recurrent all-pairs field transforms for optical flow,'' in \emph{European Conference on Computer Vision}.\hskip 1em plus 0.5em minus 0.4em\relax Springer, 2020, pp. 402--419.

\bibitem{dong2024memflow}
Q.~Dong and Y.~Fu, ``Memflow: Optical flow estimation and prediction with memory,'' in \emph{Proceedings of the IEEE/CVF Conference on Computer Vision and Pattern Recognition (CVPR)}, June 2024, pp. 19\,068--19\,078.

\bibitem{Morimitsu2024RecurrentPartialKernel}
H.~Morimitsu, X.~Zhu, X.~Ji, and X.-C. Yin, ``Recurrent partial kernel network for efficient optical flow estimation,'' \emph{Proceedings of the AAAI Conference on Artificial Intelligence}, vol.~38, no.~5, pp. 4278--4286, Mar. 2024.

\bibitem{Morimitsu2025DPFlow}
H.~Morimitsu, X.~Zhu, R.~M. Cesar, X.~Ji, and X.-C. Yin, ``Dpflow: Adaptive optical flow estimation with a dual-pyramid framework,'' in \emph{Proceedings of the IEEE/CVF Conference on Computer Vision and Pattern Recognition (CVPR)}, June 2025, pp. 17\,810--17\,820.

\bibitem{kitti2012}
A.~Geiger, P.~Lenz, C.~Stiller, and R.~Urtasun, ``Vision meets robotics: The kitti dataset,'' \emph{The International Journal of Robotics Research}, vol.~32, no.~11, pp. 1231--1237, 2013.

\bibitem{butler2012naturalistic}
D.~J. Butler, J.~Wulff, G.~B. Stanley, and M.~J. Black, ``A naturalistic open source movie for optical flow evaluation,'' in \emph{European conference on computer vision}.\hskip 1em plus 0.5em minus 0.4em\relax Springer, 2012, pp. 611--625.

\bibitem{mayer2016large}
N.~Mayer, E.~Ilg, P.~Hausser, P.~Fischer, D.~Cremers, A.~Dosovitskiy, and T.~Brox, ``A large dataset to train convolutional networks for disparity, optical flow, and scene flow estimation,'' in \emph{Proceedings of the IEEE conference on computer vision and pattern recognition}, 2016, pp. 4040--4048.

\bibitem{Dosovitskiy_2015_ICCV}
A.~Dosovitskiy, P.~Fischer, E.~Ilg, P.~Hausser, C.~Hazirbas, V.~Golkov, P.~van~der Smagt, D.~Cremers, and T.~Brox, ``Flownet: Learning optical flow with convolutional networks,'' in \emph{Proceedings of the IEEE International Conference on Computer Vision (ICCV)}, December 2015.

\bibitem{kondermann2016hci}
D.~Kondermann, R.~Nair, K.~Honauer, K.~Krispin, J.~Andrulis, A.~Brock, B.~Gussefeld, M.~Rahimimoghaddam, S.~Hofmann, C.~Brenner \emph{et~al.}, ``The hci benchmark suite: Stereo and flow ground truth with uncertainties for urban autonomous driving,'' in \emph{Proceedings of the IEEE Conference on Computer Vision and Pattern Recognition Workshops}, 2016, pp. 19--28.

\bibitem{kerbl3Dgaussians}
\BIBentryALTinterwordspacing
B.~Kerbl, G.~Kopanas, T.~Leimk{\"u}hler, and G.~Drettakis, ``3d gaussian splatting for real-time radiance field rendering,'' \emph{ACM Transactions on Graphics}, vol.~42, no.~4, July 2023. [Online]. Available: \url{https://repo-sam.inria.fr/fungraph/3d-gaussian-splatting/}
\BIBentrySTDinterwordspacing

\bibitem{Wu_2024_CVPR}
G.~Wu, T.~Yi, J.~Fang, L.~Xie, X.~Zhang, W.~Wei, W.~Liu, Q.~Tian, and X.~Wang, ``4d gaussian splatting for real-time dynamic scene rendering,'' in \emph{Proceedings of the IEEE/CVF Conference on Computer Vision and Pattern Recognition (CVPR)}, June 2024, pp. 20\,310--20\,320.

\bibitem{qian2024gaussianavatars}
S.~Qian, T.~Kirschstein, L.~Schoneveld, D.~Davoli, S.~Giebenhain, and M.~Nie{\ss}ner, ``Gaussianavatars: Photorealistic head avatars with rigged 3d gaussians,'' in \emph{Proceedings of the IEEE/CVF Conference on Computer Vision and Pattern Recognition}, 2024, pp. 20\,299--20\,309.

\bibitem{offtanet}
J.~Zhang, F.~Liu, and A.~Zhou, ``Off-tanet: A lightweight neural micro-expression recognizer with optical flow features and integrated attention mechanism,'' in \emph{PRICAI 2021: Trends in Artificial Intelligence}, D.~N. Pham, T.~Theeramunkong, G.~Governatori, and F.~Liu, Eds.\hskip 1em plus 0.5em minus 0.4em\relax Cham: Springer International Publishing, 2021, pp. 266--279.

\bibitem{HORN1981185}
\BIBentryALTinterwordspacing
B.~K. Horn and B.~G. Schunck, ``Determining optical flow,'' \emph{Artificial Intelligence}, vol.~17, no.~1, pp. 185--203, 1981. [Online]. Available: \url{https://www.sciencedirect.com/science/article/pii/0004370281900242}
\BIBentrySTDinterwordspacing

\bibitem{tvl1}
J.~Sánchez, E.~Llopis, and G.~Facciolo, ``Tv-l1 optical flow estimation,'' \emph{Image Processing On Line}, vol.~3, pp. 137--150, 07 2013.

\bibitem{gf}
G.~Farnebäck, ``Two-frame motion estimation based on polynomial expansion,'' vol. 2749, 06 2003, pp. 363--370.

\bibitem{liu2020learning}
L.~Liu, J.~Zhang, R.~He, Y.~Liu, Y.~Wang, Y.~Tai, D.~Luo, C.~Wang, J.~Li, and F.~Huang, ``Learning by analogy: Reliable supervision from transformations for unsupervised optical flow estimation,'' in \emph{IEEE Conference on Computer Vision and Pattern Recognition(CVPR)}, 2020.

\bibitem{yang2019vcn}
G.~Yang and D.~Ramanan, ``Volumetric correspondence networks for optical flow,'' in \emph{NeurIPS}, 2019.

\bibitem{Yin_2019_CVPR}
Z.~Yin, T.~Darrell, and F.~Yu, ``Hierarchical discrete distribution decomposition for match density estimation,'' in \emph{The IEEE Conference on Computer Vision and Pattern Recognition (CVPR)}, June 2019.

\bibitem{jiang2021learning}
S.~Jiang, D.~Campbell, Y.~Lu, H.~Li, and R.~Hartley, ``Learning to estimate hidden motions with global motion aggregation,'' in \emph{ICCV}, 2021.

\bibitem{Sun2022SKFlowLearningOptical}
S.~SUN, Y.~Chen, Y.~Zhu, G.~Guo, and G.~Li, ``Skflow: Learning optical flow with super kernels,'' in \emph{Advances in Neural Information Processing Systems}, S.~Koyejo, S.~Mohamed, A.~Agarwal, D.~Belgrave, K.~Cho, and A.~Oh, Eds., vol.~35.\hskip 1em plus 0.5em minus 0.4em\relax Curran Associates, Inc., 2022, pp. 11\,313--11\,326.

\bibitem{xu2022unifying}
H.~Xu, J.~Zhang, J.~Cai, H.~Rezatofighi, F.~Yu, D.~Tao, and A.~Geiger, ``Unifying flow, stereo and depth estimation,'' \emph{arXiv preprint arXiv:2211.05783}, 2022.

\bibitem{jahedi2024ccmr}
A.~Jahedi, M.~Luz, M.~Rivinius, and A.~Bruhn, ``Ccmr: High resolution optical flow estimation via coarse-to-fine context-guided motion reasoning,'' in \emph{Proceedings of the IEEE/CVF Winter Conference on Applications of Computer Vision}, 2024, pp. 6899--6908.

\bibitem{wang2024sea}
Y.~Wang, L.~Lipson, and J.~Deng, ``Sea-raft: Simple, efficient, accurate raft for optical flow,'' \emph{arXiv preprint arXiv:2405.14793}, 2024.

\bibitem{xu2022gmflow}
S.~Zhao, L.~Zhao, Z.~Zhang, E.~Zhou, and D.~Metaxas, ``Global matching with overlapping attention for optical flow estimation,'' in \emph{Proceedings of the IEEE/CVF Conference on Computer Vision and Pattern Recognition}, 2022.

\bibitem{Morimitsu2024RAPIDFlowRecurrentAdaptable}
H.~Morimitsu, X.~Zhu, R.~M. Cesar-Jr, X.~Ji, and X.-C. Yin, ``{RAPIDFlow}: {Recurrent Adaptable Pyramids with Iterative Decoding} for efficient optical flow estimation,'' in \emph{International Conference on Robotics and Automation}, 2024.

\bibitem{Mehl2023_Spring}
L.~Mehl, J.~Schmalfuss, A.~Jahedi, Y.~Nalivayko, and A.~Bruhn, ``Spring: A high-resolution high-detail dataset and benchmark for scene flow, optical flow and stereo,'' in \emph{Proceedings of the IEEE/CVF Conference on Computer Vision and Pattern Recognition (CVPR)}, June 2023, pp. 4981--4991.

\bibitem{ssa}
M.~Alkaddour, U.~Tariq, and A.~Dhall, ``Self-supervised approach for facial movement based optical flow,'' \emph{IEEE Transactions on Affective Computing}, vol.~13, no.~4, pp. 2071--2085, 2022.

\bibitem{peng2023facial}
Z.~Peng, B.~Jiang, H.~Xu, W.~Feng, and J.~Zhang, ``Facial optical flow estimation via neural non-rigid registration,'' \emph{Computational Visual Media}, vol.~9, no.~1, pp. 109--122, 2023.

\bibitem{kemmou5457671hybrid}
A.~KEMMOU, A.~EL~MAKRANI, I.~EL~AZAMI, and M.~AABIDI, ``Hybrid convolutional-transformer generative network for automatic facial expression recognition under partial occlusions,'' \emph{Available at SSRN 5457671}.

\bibitem{kirschstein2023nersemble}
\BIBentryALTinterwordspacing
T.~Kirschstein, S.~Qian, S.~Giebenhain, T.~Walter, and M.~Nie\ss{}ner, ``Nersemble: Multi-view radiance field reconstruction of human heads,'' \emph{ACM Trans. Graph.}, vol.~42, no.~4, jul 2023. [Online]. Available: \url{https://doi.org/10.1145/3592455}
\BIBentrySTDinterwordspacing

\bibitem{qian2024vhap}
\BIBentryALTinterwordspacing
S.~Qian, ``Vhap: Versatile head alignment with adaptive appearance priors,'' sep 2024. [Online]. Available: \url{https://github.com/ShenhanQian/VHAP}
\BIBentrySTDinterwordspacing

\bibitem{FLAME:SiggraphAsia2017}
\BIBentryALTinterwordspacing
T.~Li, T.~Bolkart, M.~J. Black, H.~Li, and J.~Romero, ``Learning a model of facial shape and expression from {4D} scans,'' \emph{ACM Transactions on Graphics, (Proc. SIGGRAPH Asia)}, vol.~36, no.~6, pp. 194:1--194:17, 2017. [Online]. Available: \url{https://doi.org/10.1145/3130800.3130813}
\BIBentrySTDinterwordspacing

\bibitem{davison2016samm}
A.~K. Davison, C.~Lansley, N.~Costen, K.~Tan, and M.~H. Yap, ``Samm: A spontaneous micro-facial movement dataset,'' \emph{IEEE transactions on affective computing}, vol.~9, no.~1, pp. 116--129, 2016.

\bibitem{yan2014casme}
W.-J. Yan, X.~Li, S.-J. Wang, G.~Zhao, Y.-J. Liu, Y.-H. Chen, and X.~Fu, ``Casme ii: An improved spontaneous micro-expression database and the baseline evaluation,'' \emph{PloS one}, vol.~9, no.~1, p. e86041, 2014.

\bibitem{li2013spontaneous}
X.~Li, T.~Pfister, X.~Huang, G.~Zhao, and M.~Pietik{\"a}inen, ``A spontaneous micro-expression database: Inducement, collection and baseline,'' in \emph{2013 10th IEEE International Conference and Workshops on Automatic face and gesture recognition (fg)}.\hskip 1em plus 0.5em minus 0.4em\relax IEEE, 2013, pp. 1--6.

\bibitem{Quang2019Capsulenet}
N.~V. {Quang}, J.~{Chun}, and T.~{Tokuyama}, ``Capsulenet for micro-expression recognition,'' in \emph{2019 14th IEEE International Conference on Automatic Face Gesture Recognition (FG 2019)}, 2019, pp. 1--7.

\bibitem{ilg2018occlusions}
E.~Ilg, T.~Saikia, M.~Keuper, and T.~Brox, ``Occlusions, motion and depth boundaries with a generic network for disparity, optical flow or scene flow estimation,'' in \emph{Proceedings of the European conference on computer vision (ECCV)}, 2018, pp. 614--630.

\end{thebibliography}


 





\end{document}